# The Importance of Models in Data Analysis with Small Human Movement Datasets - Inspirations from Neurorobotics Applied to Posture Control of Humanoids and Humans

Vittorio Lippi[1][a], Christoph Maurer[1][b] and Thomas Mergner[1][c]
[1]Neurological University Clinic, University of Freiburg, Freiburg im Breisgau, Germany
vittorio.lippi@uniklinik-freiburg.de, thomas.mergner@uniklinik-freiburg.de, christoph.maurer@uniklinik-freiburg.de

Abstract: Machine learning has shown impressive improvements recently, thanks especially to the results shown in deep learning applications. Besides important advancements in the theory, such improvements have been associated with an increment in the complexity of the used models (i.e. the numbers of neurons and connections in neural networks). Bigger models are possible given the amount of data used in the training process is increased accordingly. In medical applications, however, the size of datasets is often limited by the availability of human subjects and the effort required to perform human experiments. This position paper proposes the integration of bioinspired models with machine learning.

## 1. INTRODUCTION

During the last decade, there have been great improvements in machine learning applications, in the sense that the machine learning systems got more powerful and accurate. This improvement is associated with a resurgence of the use of neural networks, in particular of deep learning. As shown in Fig 1, the size of the neural networks has increased in the order of magnitudes during the last 40 years as has the number of samples used for the training. A massive dataset of training samples is not always available, however. In the case of data from human experiments, the reason for the difficulty in getting a huge amount of data lies in the effort required to perform the experiments and in the fact that human data are often described by a large number of relevant features; in some cases, there are more features than samples (Hastie & Tibshirani, 2004). For this reason, when working with human data, regularization is of primary importance. Deep learning systems are finding application in the analysis of human movements (Abdu-Aguye & Gomaa, 2019b, 2019a) and, while the results are promising, the field is still at the beginning and hence the possibilities are still to be fully explored. In this position paper, we will present examples that show the advantage of integrating models in the analysis of human experiments. The particular case of human and humanoid posture control is presented and some examples will be discussed. The application of ML to human posture control analysis is already a research topic, for example to design diagnostic tools in a clinical setup (Costa et al., 2016). The issue will be shown from the point of view of both the analysis of human data and the control of humanoid robots' balance. Modern research on human and humanoid posture control already uses mathematical models (Alexandrov et al., 2017; Boonstra et al., 2014; Engelhart et al., 2014; Goodworth & Peterka, 2018; Mergner, 2010; Pasma et al., 2014; van Asseldonk et al., 2006; H van der Kooij et al., 2007; Herman van der Kooij et al., 2005). The presented models are designed to describe, and in some cases predict, human behavior in specific experiments, and they incorporate hypotheses about neural movement control and empirical findings. It comes natural when applying machine learning to also try to integrate the knowledge represented by such models with the adaptability of the learning systems. The examples presented in the following will try to cover different applications (i.e. classification, control, and system identification) and show the advantages of the integration of modelling and learning. The methods used in the two examples are on-line linear regression and deep learning (convolutional neural network); they are presented not with the intention to compare different ML methods but to show how posture control models can be integrated in different set-ups.

---

[a] https://orcid.org/0000-0001-5520-8974
[b] https://orcid.org/0000-0001-9050-279X
[c] https://orcid.org/0000-0001-7231-164X

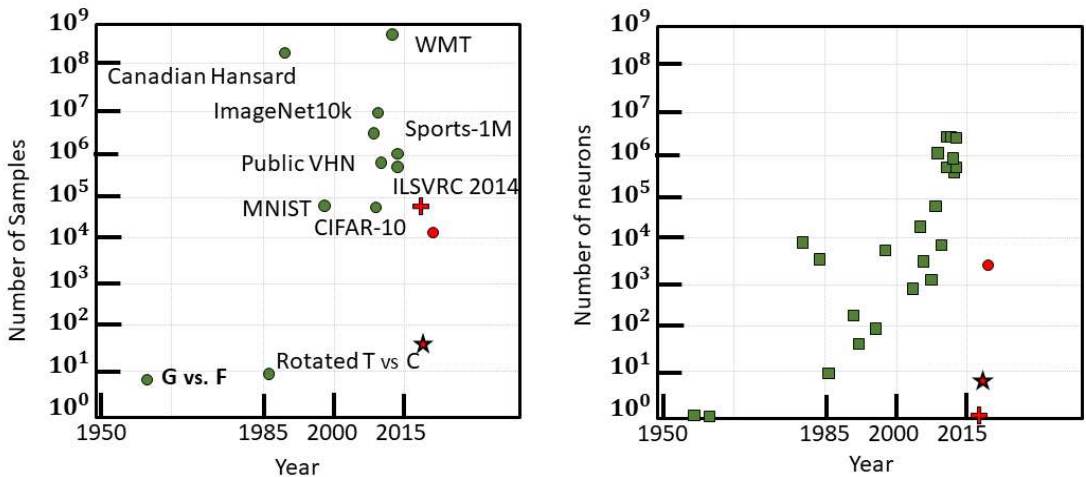

Fig 1. Increase of dataset sizes and numbers of neurons of the neural network through the last 70 years in comparison with recent posture control and balance applications. On the left (green dots) the number of samples in several datasets used in machine learning applications, on the right (green squares) the number of neurons in neural networks developed during the years (from the *Perceptron* to *GoogleNet*). The graphs are adapted from (Goodfellow et al., 2016) where the complete list of NN architectures presented in the figure is available. The red marks represent the number of samples or subjects used in recent applications (on the left plot) and the size of the neural network in the respective solution (right plot). Specifically, the star represents (Jafari et al., 2019), the cross Lippi (2018), and the dot (Lippi et al., 2020). It is evident how the three posture control examples rely on smaller datasets and smaller architectures compared to the possibilities of deep learning at the state of the art.

## 2. EXAMPLES

## 2.1 The disturbance identification and compensation (DEC) model for posture control

The examples presented in this section will make use of a bio-inspired posture control model, the DEC (Mergner et al., 2003). A brief description of the model is provided as an introduction to the following examples for a more in-depth description see Lippi & Mergner (2017), where the DEC is implemented as a modular control system for humanoid robots. The DEC control is designed to a describe how human postural control mechanisms interact with movement execution control. A schema of the DEC control is shown in Fig. 2 (top), The components of the control are: (A) A servo control loop for each degree of freedom. The controller is a PD controller, or PID in some implementations (the block "C" in Fig 2.). The controlled variable consists either of the joint angle, the orientation in space of the above joint, or the orientation in space of the centre of mass of the whole body above the controlled joint. The control is implemented in a modular way, and each module performs sensor fusion and control. (B) Multisensory estimation of external disturbances, i.e. rotation and translation of the supporting link or support, contact forces, and field forces such as gravity. The disturbance estimates are fed into the servo so that the joint torque compensates on-line for the disturbances while executing the desired movements.

The disturbance compensation mechanism allows the system to use a low loop gain and thus stable control in face of neural time delays, or, in case of humanoid control, of delays due to signal transmission or low sample rate (Ott et al., 2016). The reference input to each module determines its postural function, e.g. maintaining a given orientation of the supported link (either in space or with respect to the supporting link), or maintaining the COM above its supporting joint. Modules exchange information with neighbouring modules, i.e. those mechanically interconnected.

## 2.2 Online learning for the posture control of the Lucy robot

Small human/humanoid datasets may suffice to use linear learning systems. As an example, our previous work (Lippi, 2018) shows how the nonlinear DEC model can be integrated with a linear learning system to make it capable of controlling human posture control. The challenge here was represented by the closed-loop nature of the posture control, i.e. by the fact that the body is intrinsically unstable and the control is always active. The machine learning process is then based on data that are influenced by the use of the learned predictor itself. Therefore, an online training approach was proposed. It improved the control of the body sway without endangering control stability. In Fig. 2 the structure of the bioinspired predictor is shown. The ML model was a rather simple linear model, implemented in a way so that it could learn incrementally as the robot was balancing.

In particular, previously and here the learning systems are trained to predict the COM sway $\alpha_{BS}$, with an anticipation of $T_{pred} = 70$ ms. The inputs taken into account are the previous sensory-based values for the body sway angle $\alpha_{bs}$ and the reference value $y_i = \bar{\alpha}_{BS}$ (sampled at previous steps). Every 10 ms an input vector is constructed using delayed versions of the input signals:

$$x_i = [\alpha_{bs}(\tau) \quad \alpha_{bs}(\tau - \Delta t) \quad \alpha_{bs}(\tau - 2\Delta t) \quad \bar{\alpha}_{BS}(\tau - \Delta t) \quad \bar{\alpha}_{BS}(\tau - \Delta t) \quad \bar{\alpha}_{BS}(\tau - 2\Delta t)]$$

where $\tau = t_i - T_{pred}$ and $\Delta t$ is set to 64 ms. The predictor has the structure of an affine application, where the parameter to be learned are the elements of the transformation matrix. Specifically, the disturbance to be predicted at the time $i$, $y_i$ can be arranged in a vector of target values $Y = [y_1 \ y_2 \ \cdots \ y_n]$, and the observed input is integrated into the matrix

$$X = \begin{bmatrix} x_1^T & x_2^T & \cdots & x_n^T \\ 1 & 1 & & 1 \end{bmatrix} \quad (1)$$

The weight matrix is computed as $W = YX^{\dagger}$, using the pseudoinverse operation that can be implemented on-line. The values used to build X and Y are affected by the prediction, as shown in Fig. 2.

The use of the real robotic platform Lucy, with real noisy sensors, helped to evaluate the hypothesis about predictions in a real-world setup. The robustness of the system was tested including an additional delay in the loop. The prediction system allows the system to stand with a delay of 60 ms, while the system without prediction becomes unstable at 10 ms. The prediction system was compared with a Smith predictor (that is based just on the model of the system) and, as result, proved to produce a better performance.

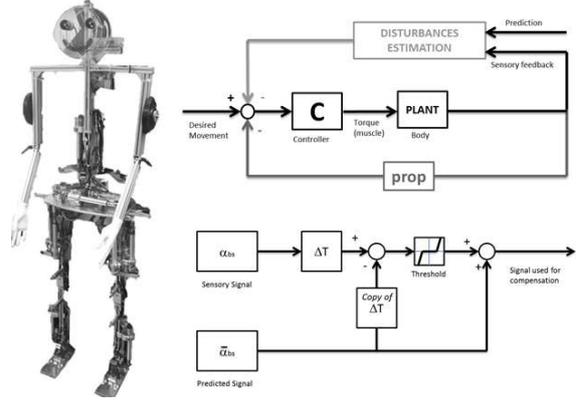

Fig. 2. Integration of the prediction system based on a linear learning model. On the left the Lucy robot, a humanoid with 14 DoF, where the system was tested. The schema above shows how the DEC control integrates disturbances estimation. In this specific case, the predicted effect is the gravity torque. The "prop" block represents the proprioceptive feedback, based on joint angles, while "disturbances estimation" is implemented through a sensor fusion integrating proprioception and vestibular (IMU) input. The prediction is compared with the measured value as shown in the schema below: The threshold function has the effect that, when the prediction and the sensor value are similar, the prediction is used, while the sensed value is used when the difference between the two is large. This approach resembles a Smith predictor and the way the efference copy mechanism is used in modeling human behavior.

## 2.3 How models can benefit from Machine Learning (ML): System identification with CNN

A previous work (Lippi et al., 2020) presented a method for posture control parameter identification based on CNN. It represents an example of how ML can provide a tool for modelling, exploiting the knowledge of the posture control system in the form of a parametric model; the CNN identifies the parameters of such a model.

Human posture control exhibits nonlinearities such as deadbands and gain non-linearities. Nonlinear models are more complex to be fitted on human data than linear models and, in the general case, expensive iterative procedures need to be used. This issue brought us to the idea to identify the parameters of a nonlinear bio-inspired posture control system, the DEC model using ML. The advantage lies here in the fact that using the trained network is almost immediate, whereas training the CNN would be more computationally expensive.

The training set was produced with parameters from uniform distributions, filtered with the constraint that they would produce a stable simulation. The number of samples can be as large as needed, being here produced through a simulation. In order to obtain more human-like examples, the data-set was enriched with samples of larger body sways. In the future, the CNN can also be tested *a posteriori* by comparing the distribution of the parameters it produces for the validation set with the expected distribution for the real data. This can help in choosing hyperparameters as shown in previous works (Sforza et al., 2011; Sforza & Lippi, 2013). Fig. 3 schematically summarizes the pipeline of the work. The input of the network is a 2-channel picture, representing the modulus and the phase of the fft of the body sway computed on time windows (in Fig.3 the two channels are visualized as "green" and "blue"). Because of its architecture, i.e. training the same weights on different parts of the image, the CNN is able to recognize patterns translated in time and in frequency. While the invariance in time has the obvious advantage of making the recognition of a specific motion feature independent from its onset in the input signal, the invariance in frequency has no obvious physical interpretation. The SIP model proved to be suitable to describe the analysed posture control scenario, this even in the sub-optimal case of identifying the control parameter of the ankle joint in a DIP model.

## 3. CONCLUSIONS AND FUTURE WORK

This position paper gives two examples of the use of posture control models in learning. The examples suggest that the modeling can be useful in reducing the number of features used with the ML algorithm, simplifying the complexity of the ML system required to perform the task, or increasing the number of training samples using simulations to produce artificial data.

The identification of posture control model parameters can be applied to the benchmarking of humanoid robots (Lippi et al., 2019; Torricelli et al., 2020) and to the analysis of clinical data (Exarchos et al., 2015).

From the point of view of control applications, synergies between machine learning and posture control can find applications in the control of wearable robots. Fig. 5. shows an example presenting the hypothetical structure of the control system for a full-body exoskeleton. The actuated ankle joint and the fact that the robot's geometry prevents the user from having direct contact with the support surface implies that the robot has to balance by itself. The balance and posture control issues specific to legged humanoids apply also to wearable robots. This implies the complication of physical interactions between the robot and the human. The figure provides a map of possible applications of the ML approaches presented in the examples (Section 2) for the components of the exoskeleton control.

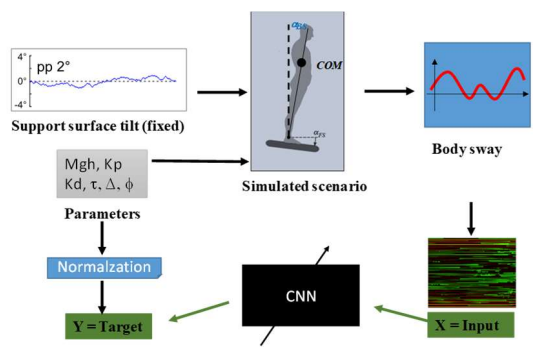

Fig.3. The pipeline of the learning problem is presented in Lippi et al.( 2020). The simulated scenario represents a subject standing on a tilting support surface. The tilt profile is a pseudo-random ternary sequence (PRTS) function for all the simulations. The parameters of the simulations are generated randomly and the output of the simulation is the profile of the body COM sway. The training process, aiming to identify the parameters, "reverses" the relationship between the data: body sway, here transformed into a picture, is the input, and the parameters, centred around the mean and divided by the variance of the training set ('Normalization' block) are the target output. The identification is formally a regression problem.

Besides posture control and balance, a wearable robot poses issues that have not been covered by the presented examples and can still be solved with proper integration of models and ML. Specifically, a transparent transfer of voluntary movements between the user and the robot requires the mapping of trajectories between different kinematic structures, even if the user's joints are not necessarily coincident with those of the robot (Godoy et al., 2018; Lee et al., 2018). Machine learning techniques provide means to also solve such problems (see for example (Makondo et al., 2015). Learning trajectories and libraries of trajectories associated with tasks, e.g. gait, can be achieved by exploiting models for movement representation such as movement primitives (Paraschos et al., 2013; Schaal et al., 2005; Schaal, 2006) and the algorithms to generalize and transfer them. For tasks such as manipulations, where reaction forces may reasonably be more important than the trajectories themselves, models representing the stiffness of the robot (e.g. Calinon, 2016; Calinon et al., 2007) or specifications of the particular mechanical variables (torques, velocity positions, etc.) involved in the task can be used (Deimel, 2019b, 2019a). In all these cases the models have a powerful regularization effect, in that a model of human motor behavior can be learned from a few samples, or even just a single sample (Schaal, 2006).

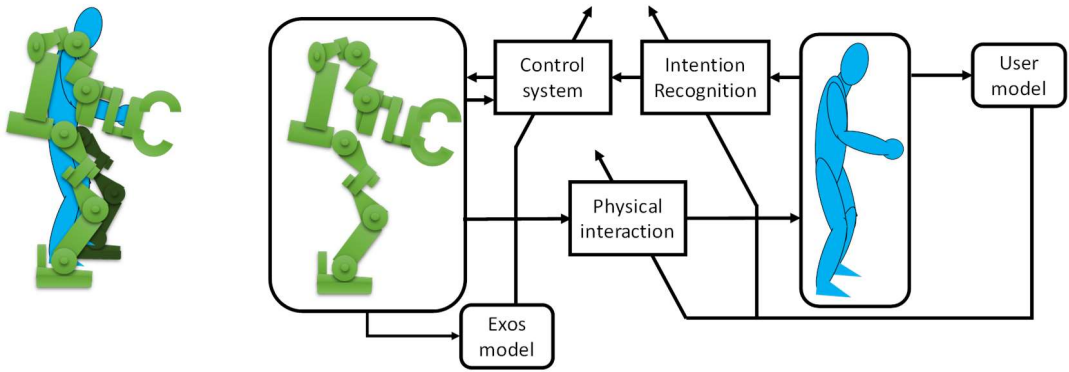

Fig 5. Hypothetical configuration of a user wearing a full-body exoskeleton. The block diagram shows the two mechanical components of the system (robot and user) as two separate blocks to highlight how their interaction is mediated by control systems that can be tuned using machine learning and thereby can benefit from the integration of modeling. The robot model "exos model" can integrate a learning process like the one presented in (Lippi, 2018). The control system parameters can be tuned accordingly. The haptic feedback that the robot here produces, "physical interaction" block, could be designed on the basis of human sensor fusion in order to map the behavior of the robot to match the perception of the user (for example, the robot should be in equilibrium when the user perceives himself as being in equilibrium). For this purpose, using a model of the user's posture control, the "user model", can be beneficial. On the other hand, such a model can also be used to anticipate the user's movements in the block "intention recognition", which is used to provide commands to the control system of the robot. Both the "Exos model" and the "User model" can be identified by means of machine learning (Lippi et al., 2020).

The topic of reinforcement learning (RL) has not been considered in specific examples. RL is a popular way to solve problems where a measure of success can be formalized (e.g, body sway amplitude, number of falls of a robot) but the desired output may not be explicitly available. An example can be the closed-loop control in section 2.2 and in general the problem of humanoid balance (e.g. in Phaniteja et al., 2018; Vuga et al., 2013; Yang et al., 2017). As RL relies on the exploration of a space of possible control policies it can benefit substantially from training in simulations (where making a mistake is not expensive) and hence it can exploit posture control models.

Overall, we contend that the proposed examples suggest that knowledge of human behaviour models (be they bio-inspired or just descriptive of a given outcome) as well as models of human sensorimotor functions are crucial for the analysis of human behavioural data. The models may provide powerful tools for the control of humanoid robots. Both the functionality of the bio-inspired models and the modern ML techniques will benefit from being mutually integrated.

# ACKNOWLEDGEMENTS

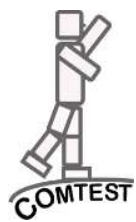

This work is supported by the project COMTEST, a sub-project of EUROBENCH (European Robotic Framework for Bipedal Locomotion Benchmarking, www.eurobench2020.eu) funded by the H2020 Topic ICT 27-2017 under grant agreement number 779963.